# An Automatic Reader of Identity Documents

Filippo Attivissimo, *Member, IEEE*, Nicola Giaquinto, *Member, IEEE*,
Marco Scarpetta, *Student Member, IEEE* and Maurizio Spadavecchia *Member, IEEE*

*Abstract*—Identity documents automatic reading and verification is an appealing technology for nowadays service industry, since this task is still mostly performed manually, leading to waste of economic and time resources. In this paper the prototype of a novel automatic reading system of identity documents is presented. The system has been thought to extract data of the main Italian identity documents from photographs of acceptable quality, like those usually required to online subscribers of various services. The document is first localized inside the photo, and then classified; finally, text recognition is executed. A synthetic dataset has been used, both for neural networks training, and for performance evaluation of the system. The synthetic dataset avoided privacy issues linked to the use of real photos of real documents, which will be used, instead, for future developments of the system.

I. INTRODUCTION

Many works have tried to address the problem of reading identity documents from photos, since this is a topic of interest for current service companies and administrative offices.

In a typical scenario, the online subscription of a contract (e.g. for telephone services, banking, etc.) requires that the user sends through a scanned identity document. A condition for the validity of the contract is that the data in the document are the same provided by the subscriber in the subscription form, and that the document is not expired. Checking that the scanned document meets the required condition is mandatory, and must be done by humans. Since it is a time-consuming and boring activity, it is costly and particularly prone to human errors. The problem is important for companies that must manage big amounts of contracts, with a wide variety of subscribers (not always inclined to use new technologies), and in some activities of public administrations (e.g., public selections with a large number of candidates, etc.)

In the future this problem will certainly be overcome, or at least radically transformed, by the advent of new solutions for remote personal identification. But, for at least some years, the human check of large amounts of scanned documents will remain the only possibility.

On the market there are solutions for the automatic reading of documents, such as BlinkID [1]. They work very well, but are based on the continuous acquisition of images of the document, which goes on until a reliable enough extraction of the data is obtained. Such solutions can be used, therefore, only together with a specific app for smartphone or similar devices: the app performs the requested continuous acquisition. Extracting data from a single photo or a single scanned copy of a document, obtained e.g. via email, is a different and much more difficult problem. Therefore, in this paper, the design of an automatic reader which only needs a simple photo to extract data from ID documents is addressed.

The first difficulty in elaborating a single photo of a document, produced by the average subscriber with her/his own smartphone (or copier, or scanner) is the localization of the document in the image, i.e. the separation of the document and the background. These two elements of the image may be hardly distinguishable for a number of reasons, including poor contrast, similar color tones, etc. A different, but often connected problem, is the classification of the document (passport, driving license, identity card, etc.), among a set of possible ones.

Different approaches have been used for solving these problems. In [2] a computer vision technique, based on visual saliency (instead of edge or contour detection) is used to locate identity documents inside a photo or video frame, without any prior knowledge about the document. Another approach is used in [3], where the document is simultaneously located and classified, by extracting a set of keypoints from the image, and comparing them with a previously built database. A similar technique is used in [4], but for classification only purposes. In this case, the extracted features are used to create a histogram of the photo, which is then classified. In [5], different classification technique are applied to identity documents. Complete systems have also been proposed [6]–[8], which make use of artificial intelligence algorithm for text recognition.

In short, our proposed system localizes the document making use of color information to distinguish the document region from the background region. Once the document has been localized, it is classified using a convolutional neural network (CNN) model. Textual information is finally read from the document using a single-line text recognition model. This is possible since the position of text regions in each document type is fixed.

II. DESCRIPTION OF THE SYSTEM

The system was designed to read data from the main Italian identity documents, that is:

- paper identity card;

The Authors are with Politecnico di Bari, Department of Electrical and Information Engineering (DEI), via Orabona 4, I-70125 Bari (BA), Italy.
Contact information:
F. Attivissimo - email filippo.attivissimo@poliba.it
N. Giaquinto - email nicola.giaquinto@poliba.it
M. Scarpetta (corresponding author) – email marco.scarpetta@poliba.it, phone +39 080 596 3583
M. Spadavecchia – email maurizio.spadavecchia@poliba.it

- electronic identity card;
- driving license;
- health insurance card;
- passport.

Both front and back side are considered for each of them, except the passport. The system takes photos of identity documents as input data, and process them through a series of stages, described in the following subsections, to finally obtain a digital version of the personal data contained in the document.

*A. Vertices detection*

The first step in identity documents reading is the detection of the document position inside the input image, which can be described by its four vertices. We have accomplished this operation using a simple algorithm, alternative to others based on edge and contour detection. The algorithm works by adjusting iteratively the vertices of the documents, using data of pixels sampled in an "outer" region.

The outer region is the outside of a rectangle with edges at 7% and 93% of the image (dashed white line in Fig. 1). Assuming that the document is approximatively in the center of the image, the region includes only pixels belonging to the background. The starting position of the vertices are those of a rectangle with edges at 30% and 70% of the image width and height (dashed blue line in Fig. 1). It can be assumed that the true vertices are outside this rectangle.

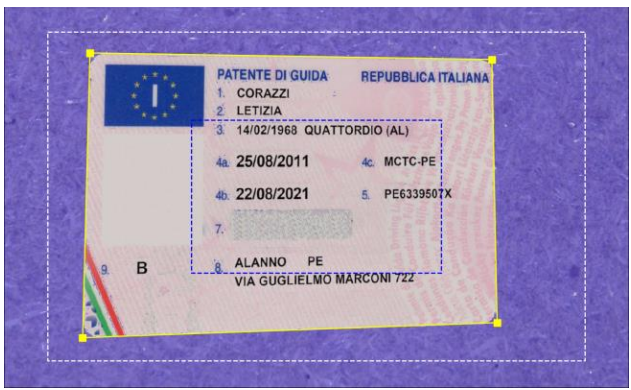

Fig. 1. Identification of the document vertices. The starting document vertices are those of the blue dashed rectangle. The "outer" region, composed by pixels belonging to the background, is outside the yellow dashed line. The final document vertices are the yellow points connected by continuous lines. Note: data in the document are fictitious.

The algorithm works as follows. A certain number of pixels ($L = 100$ in our practical implementation) are sampled from the outer region. For each of them, all the pixels in the image with "similar" color are selected, where the "similarity" is decided on the basis of the Euclidean distance between RGB colors, and of an automatically adjusted threshold. Then, the goodness of the candidate vertices of the document is measured, using the function:

$$c = \sum_{i=1}^{N}\sum_{j=1}^{M} a_{ij} \cdot \overline{b_{ij}} + d \sum_{i=1}^{N}\sum_{j=1}^{M} \overline{a_{ij}} \cdot b_{ij} \quad (1)$$

where $M$ and $N$ are the width and height of the image, and, for the pixel $(i, j)$:

- $a_{ij} = 1$ if the pixel is selected, $a_{ij} = 0$ otherwise;
- $b_{ij}$ is 1 if the pixel is inside the vertices-defined region, and 0 if it is outside;
- $\bar{a}$ and $\bar{b}$ are the logical negation of $a$ and $b$;
- $d$ is a weight factor, empirically set to $d = 1.5$.

The first term is higher if there are many pixels that are similar to the background ("selected"), and are outside the vertices-defined region; the second term is higher if there are many pixels different from the background, and are included in the vertices-defined region. Therefore, the higher is $c$, the better are the candidate vertices. The algorithms move outwards the candidate vertices, until a maximum of $c$ is found. Here is an outline of the algorithm.

1) Define $L = 100, T = 25, d = 1.5$
2) Define the outer region and the candidate vertices, defining the document region
3) Sample $L$ pixels from the outer region
4) For $k = 1: L$
   a) $T(k) = T$ is the starting threshold value for the $k$-th pixel
   b) Select all pixels in the image whose color is distant less than $T(k)$
   c) While the selected pixels which are in the document region are more than 0.01% of total pixels:
      i) Decrease the threshold value: $T(k) = T(k) - 1$
      ii) Select all pixels in the image whose color is distant less than $T(k)$
5) Move outwards the vertices, until a maximum of (1) is found.

An example of vertices detection is depicted in Fig. 1, where selected background pixels are highlighted in blue and the computed document vertices are drawn in yellow.

*B. Document classification*

The following step consists in identifying the document type from the nine possible choices. An image of the document is extracted from the main photo applying a perspective transformation to make it squared with 200 pixels per side. The image is then input to a convolutional neural network model designed for classification. The model was designed following a classical scheme for classification tasks [9]. The input image is first processed through a stack of $N_b$ blocks, each composed of a convolutional layer followed by a Max Pooling layer. The former operates a convolution of the image with a set of learned filters, which serve to extract features from the image, while the latter performs a subsampling operation which filters out small translations and deformations of the image. Results of the last convolution and pooling operation are then flattened and sent to the output layer, that is composed of *softmax* units. The output of the last layer is thus an array of probability scores for each document class and the document class with maximum probability is chosen as result of the classification operation [10].

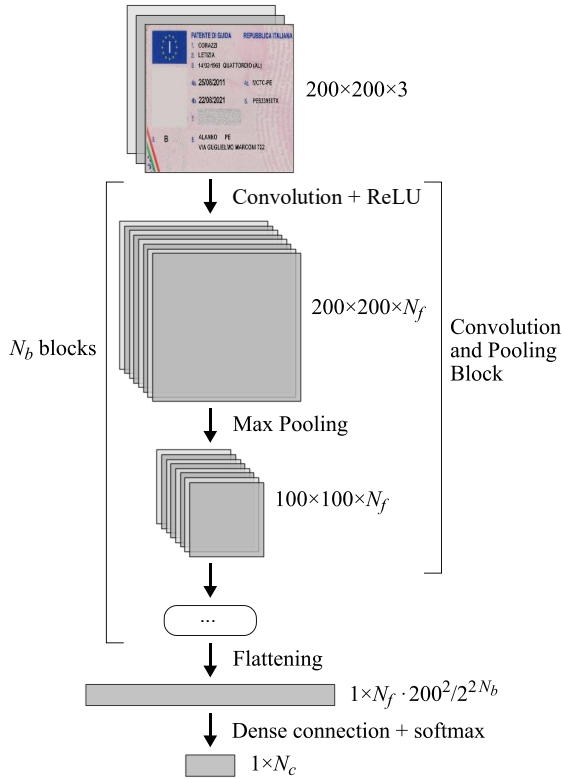

Fig. 2. Diagram of the neural network used for classification

The diagram of the classification model is depicted in Fig. 2. The input image has 3 channels (RGB) of size 200x200, which are processed by the $N_f$ filters of the first convolutional stage to produce $N_f$ feature maps of the same size. The non-linear ReLU activation function was used for the convolutional layer, as it has been showed to improve performance of deep convolutional neural networks for classification [11]. The spatial resolution of the feature maps is then reduced by the Max Pooling layer, which was chosen to have 2x2 size and strides. Output feature maps have thus half the dimensions of input feature maps.

The parameters of the model were tuned to improve the classification accuracy. The training of the model was executed using a synthetic dataset built as described in Subsection III.B. The accuracy performance of the model was evaluated using a small dataset derived from photos and scans of real identity documents. The original images were manually cropped to contain only the documents and scaled to be 200x200. The number of samples of the dataset was increased introducing modified versions of the images, that were obtained applying a light perspective transformation and randomly adjusting contrast and brightness. After the augmentation procedure the validation dataset counted 319 samples.

The results of the training for different values of the parameters $N_c$ and $N_f$ are reported in Table I. As it can be seen, the validation accuracy is 100% for every considered configuration, thus the model was chosen based on the validation loss parameter.

TABLE I
ACCURACY PERFORMANCE OF DIFFERENT CONFIGURATIONS OF THE CLASSIFICATION MODEL

| $N_b$ | $N_f$ | Total number of parameters | Validation accuracy | Validation loss | Prediction time (ms) |
|---|---|---|---|---|---|
| 1 | 8 | 0.72M | 100% | 0.0066 | 21.7 |
| 1 | 16 | 1.4M | 100% | 0.0076 | 25.1 |
| **2** | **8** | **0.18M** | **100%** | **0.0027** | **26.9** |
| 2 | 16 | 0.37M | 100% | 0.0050 | 36.7 |
| 3 | 8 | 49k | 100% | 0.015 | 29.3 |
| 3 | 16 | 0.10M | 100% | 0.0081 | 38.9 |

This parameter was computed as the mean cross-entropy of the classification results. The cross-entropy of each result was computed as:

$$CE = \sum_{i=1}^{N_c} -y_i \cdot \log p_i$$

where $N_c$ is the total number of classes (9 in this case), $y$ is the one-hot encoded true class and $p_i$ are the computed probabilities for each class (output of the softmax layer). The chosen parameters, highlighted in Table I, are $N_b = 2$ and $N_f = 8$.

*C. Text recognition*

The text recognition is the final task of the system. It proved to be very challenging, due to complex backgrounds used in identity documents. State-of-the-art OCR solutions, i.e. Tesseract [12], produced poor results since they are designed for text recognition of scanned-quality structured documents, while text regions in identity documents are sparse and have a complex background. We used therefore a recently proposed neural network model [13] for text recognition, which was shown to achieve high accuracy results on different datasets. It is a generic text recognition system, which can be used for different tasks: handwritten recognition, OCR, scene text recognition.

In order to execute text recognition, text regions need to be extracted from the photo of the document. To do this, a new image is first extracted from the original photograph and then transformed, to have the same aspect ratio of the document type which resulted from classification. A mask, that was built for each type of document analyzing real samples, is therefore applied to the document image in order to retrieve the regions containing textual information. The text recognition is executed on these images, which are resized to have 40 pixels height, while maintaining their aspect ratios. According to the naming scheme of [13] we used the following parameters for the text recognition model: $n = 8, c_1 = 64, c_2 = 256$.

III. EXPERIMENTAL RESULTS

Performance of the system was evaluated by applying the entire recognition procedure to a synthetically-generated dataset of documents photos. For privacy reason, indeed, no public dataset of identity documents is available and private ones are hard to retrieve. Using a synthetic dataset is anyway advantageous for benchmarking the system during its development, since it is completely labeled, and the number of samples can be indefinitely increased. Synthetic photos were created introducing defects and degradations typical of actual photos.

*A. Synthetic dataset generation*

The starting point for the dataset generation was a set of templates of the nine types of identity documents obtained from scanned images and a database of dummy personal data. Dummy data was generated as follows:

- Surnames were randomly picked from [14];
- Names were randomly picked from [15];
- Places of birth were randomly picked from [16];
- Home addresses were randomly picked from [17];
- Birthdates and release/expiration dates were randomly computed to look realistic;
- Fiscal codes were calculated using algorithm in [18];
- Identification numbers and other fields of identity documents were randomly created using format in real documents, inserting previously created data when needed (like in *machine readable zones* of electronic identity card and passport).

Templates were therefore filled with personal data, trying to mimic real documents appearance by using appropriate font colors, font types and defects. The document image was then placed on a background that was randomly picked from the ALOT dataset [19], which was chosen because it provides photos of complex textures in different illumination conditions, like those that can be expected as backgrounds of user-made photos. The image was finally degraded to simulate a real photo through a series of stages:

1) perspective transformation, achieved moving outward the vertices of the image by a length in the interval $[0.05l, 0.15l]$, where $l$ is the length of the shortest side of the document in pixels;

2) alteration of contrast and brightness, obtained applying the following transformation to each pixel of the image:

$$p'_{i,j,c} = \alpha \cdot p_{i,j,c} + \beta$$

where $p_{i,j,c}$ is the starting pixel value in the range $[0, 255]$, $p'_{i,j,c}$ is the new pixel value (clipped if necessary to be in the range $[0, 255]$), $i, j$ and $c$ are the row, column and channel index respectively, $\alpha$ is a number in the interval $[0.85, 1.05]$, $\beta$ is a number in the interval $[-30, 20]$;

3) noise addition;
4) jpeg compression, with 0.7 compression rate.

The generated dataset counted 1000 samples equally distributed over the document types. Examples of images of each type of document in the dataset are depicted in Figures 3-7.

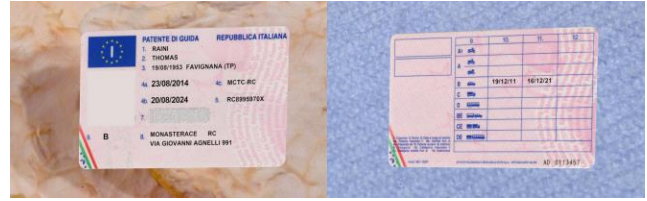

Fig. 3. Examples of synthetically-generated photos of a driving license

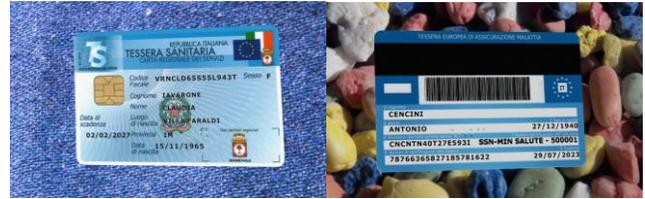

Fig. 4. Examples of synthetically-generated photos of a health insurance card

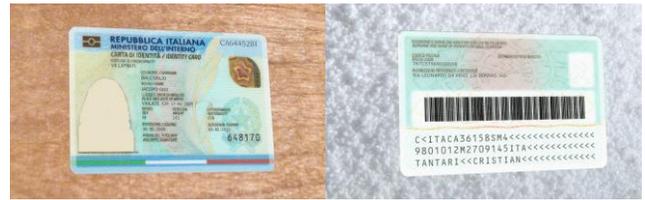

Fig. 5. Examples of synthetically-generated photos of an electronic identity card

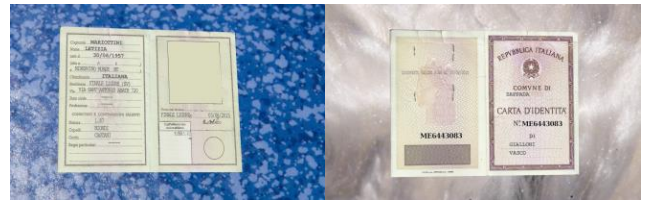

Fig. 6. Examples of synthetically-generated photos of a paper identity card

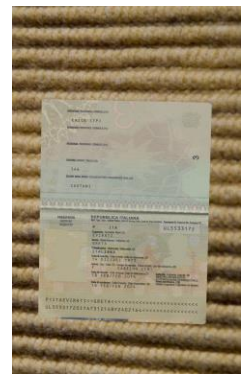

Fig. 7. Example of synthetically-generated photo of a passport

*B. Implementation details*

The whole software has been implemented in Python, using TensorFlow [20] for the artificial intelligence (AI) parts

(classification and text recognition). Synthetic datasets were also used to train the classification and the text recognition models. The synthetic dataset created for the classification model was built similarly to the main dataset but using random text strings instead of dummy personal data and producing a 200x200 image of the document. A light perspective transformation was applied to the document image to simulate errors in the vertices detection component. The text recognition model was trained on a dataset that was created extracting text regions from images of identity documents there were previously filled with random text. A light perspective transformation was also used in this case.

The classification model and the text recognition model were trained using the Adam algorithm [21] for, respectively, 239 and 201 epochs.

*C. Results*

The reading system was tested on the synthetic dataset and the output results were analyzed comparing them with the known parameters used for generating the dataset. An example of the reading procedure execution is reported in Fig. 8, where the results obtained in the several stages of the algorithm applied to a simulated photo of the front side of a driving license are depicted.

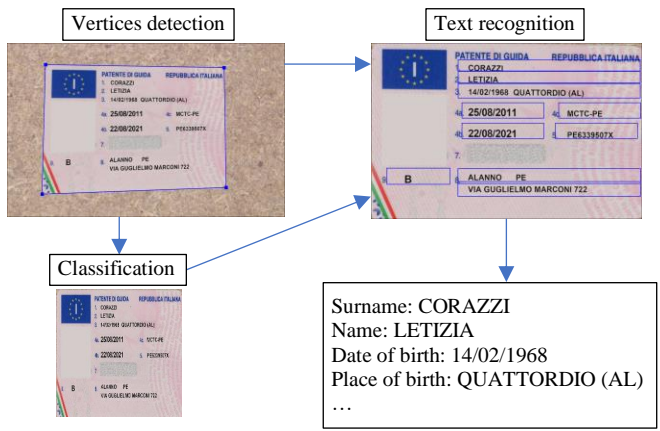

Fig. 8. Application of the reading procedure on a synthetic document photo

The vertices detection was assumed correct when the maximum displacement error of each vertex was lower than 3% of the shortest side of the document, as in this condition a tolerable distortion in the extracted document image was observed. The following stages of the procedure were obviously executed only if the vertices detection had been successful. Document vertices were correctly detected in 68.57% of cases, the classification accuracy was 100% and 92.72% of the text fields extracted from the document photo were correctly read. Errors in text recognition were more numerous for the front side of paper identity card, like it can be seen in the chart reported in Fig. 9, where the percentage of wrongly read fields over the total for each document class is represented. This was due to the higher degradation of the text for these documents, and also to the different typeface used (serif typeface). There were no errors in text recognition of back side of health insurance cards and driving licenses, where the text was more readable.

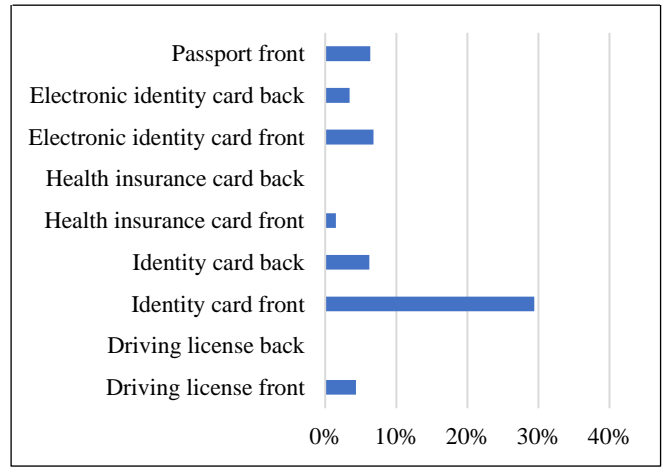

Fig. 9 Percentage of errors in text recognition

IV. CONCLUSIONS

The paper describes the first prototype of an identity documents reading system, which makes use of both classical image analysis techniques and AI techniques. Its performance has been evaluated on a synthetically-generated dataset of simulated document photos. The preliminary results presented proves that, with further development and improvements, the system can be successfully used for practical applications. The main accuracy loss was found in the vertices detection component, which can be improved, or substituted independently on the rest of the system. Accuracy of the text reading system refers to raw output, so it can be further improved exploiting additional information like the text fields format and databases of personal data.